%% file: paper.tex
\documentclass{article}

\usepackage[utf8]{inputenc} 
\usepackage[T1]{fontenc}    
\usepackage{hyperref}       
\usepackage{url}            

\usepackage{booktabs}       
\usepackage{amsfonts}       
\usepackage{nicefrac}       
\usepackage{microtype}      
\usepackage{xcolor}         
\usepackage{multirow}
\usepackage{multicol}
\usepackage{algorithm}
\usepackage[noend]{algorithmic}
\usepackage{array}
\usepackage{multirow}
\usepackage[accepted]{mlsys2025}
\usepackage{amsmath}
\usepackage{amsthm}
\usepackage{thm-restate}
\usepackage{subcaption}

\usepackage{tabularx}
\usepackage{tcolorbox}
\declaretheorem[name=Definition]{definition}

\usepackage[textsize=tiny]{todonotes}
\usepackage{amssymb}
\usepackage{listings}
\usepackage{newfloat}
\usepackage{wrapfig}
\DeclareFloatingEnvironment{listin}
\SetupFloatingEnvironment{listin}{name=Specifications}
\newenvironment{code}{\captionsetup{type=listin}}{}
\addtocounter{listin}{-1}
\newcolumntype{L}{>{\raggedleft\arraybackslash}X}

\input{macros}

\begin{document}

\twocolumn[
\mlsystitle{Specification Generation for Neural Networks in Systems}
\mlsyssetsymbol{equal}{*}

\begin{mlsysauthorlist}
\mlsysauthor{Isha Chaudhary}{to}
\mlsysauthor{Shuyi Lin}{goo}
\mlsysauthor{Cheng Tan}{goo}
\mlsysauthor{Gagandeep Singh}{to,ed}
\end{mlsysauthorlist}

\mlsysaffiliation{to}{University of Illinois, Urbana-Champaign}
\mlsysaffiliation{goo}{Northeastern University}
\mlsysaffiliation{ed}{VMWare Research}

\mlsyscorrespondingauthor{Isha Chaudhary}{isha4@illinois.edu}

\mlsyskeywords{Specifications, Computer Systems, Clustering}

\input{abs}
]
\printAffiliationsAndNotice{}
\input{intro2}

\input{background}
\input{technical}
\input{practical_technical}
\input{evaluation}

\input{relatedwork}

\input{conclusion}
\newpage

\bibliography{paper}
\bibliographystyle{plainnat}
\clearpage
\appendix
\input{appendix}
\end{document}

%% file: macros.tex
\usepackage{xspace}

\newcommand{\abrs}{{\mathcal{R}}}
\newcommand{\numabrs}{{q}}

\newcommand{\pre}{\varphi}
\newcommand{\post}{\psi}
\newcommand{\setspecs}{{\Psi}}
\newcommand{\spec}{{\mathcal{S}}}
\newcommand{\inputs}{{\mathcal{X}}}
\newcommand{\outputs}{{\mathcal{Y}}}
\newcommand{\numinputs}{m}

\newcommand{\commonbeh}{\gamma}
\newcommand{\threshcov}{\tau_{cov}}
\newcommand{\threshrep}{\tau_{rep}}
\newcommand{\threshtight}{\tau_{max}}
\newcommand{\commonbehreg}{\Gamma}
\newcommand{\tool}{SpecTRA\xspace}
\newcommand{\data}{\mathcal{D}}
\newcommand{\parts}{p}
\newcommand{\numoutputs}{k}
\newcommand{\card}{\eta}
\newcommand{\outsubset}{\Omega}

\newcommand{\support}{Sup}

\newcommand{\conf}{Conf}

%% file: abs.tex
\begin{abstract}
Specifications --- precise mathematical representations of correct domain-specific behaviors --- are crucial to guarantee the trustworthiness of computer systems. With the increasing development of neural networks as computer system components, specifications gain more importance as they can be used to regulate the behaviors of these black-box models. Traditionally, specifications are designed by domain experts based on their intuition of correct behavior. However, this is labor-intensive and hence not a scalable approach as computer system applications diversify. We hypothesize that the traditional (aka reference) algorithms that neural networks replace for higher performance can act as effective proxies for correct behaviors of the models, when available. This is because they have been used and tested for long enough to encode several aspects of the trustworthy/correct behaviors in the underlying domain. Driven by our hypothesis, we develop a novel automated framework, \tool{} (\textbf{Spec}ifications from \textbf{T}rustworthy \textbf{R}eference \textbf{A}lgorithms) to generate specifications for neural networks using references. We formulate specification generation as an optimization problem and solve it with observations of reference behaviors. \tool{} clusters similar observations into compact specifications. We present specifications generated by \tool{} for neural networks in adaptive bit rate and congestion control algorithms. Our specifications show evidence of being correct and matching intuition. Moreover, we use our specifications to show several unknown vulnerabilities of the SOTA models for computer systems. 
\end{abstract}

%% file: intro2.tex
\section{Introduction}

Neural Networks (NNs) have recently found numerous applications as integral
components in computer systems~\citep{aurora,pensieve,decima,mendis2019ithemalaccurateportablefast}.
For example, they have been applied to enhance video
streaming quality, congestion control, database query optimization, indexing,
scheduling, and various other system tasks.
Compared to traditional heuristic-based approaches,
NNs reduce development overheads
and offer improved average performance~\cite{kraska2021towards}.

Despite recent advancements, skepticism remains about the practicality of
NNs in computer systems. A key concern is that user-facing systems,
like video streaming, require high standards of performance and
reliability in dynamic environments.
Although NNs often surpass traditional methods in
average performance, they are fundamentally opaque~\citep{comet}
and lack guarantees.

Indeed, researchers have identified numerous counterintuitive behaviors
of these NNs in computer systems.
For example, \citet{verifying_learning_augmented_systems}
observed that Aurora~\cite{aurora}, a congestion control system,
in certain conditions,
would repeatedly decrease its sending rate,
ultimately reaching and maintaining the minimal rate
despite excellent network conditions.
In another example, \citet[\S6.3]{meng2020interpreting}
``debugged'' Pensieve~\cite{pensieve}, a bitrate controller for video streaming,
which systematically avoided two specific bitrates (1200 and 2850kbps);
they introduced these bitrates back to enhance performance.

Note that these observations are not unique; similar behaviors have been extensively
studied in other domains, such as vision~\citep{szegedy2014intriguingpropertiesneuralnetworks,wireless_attack,athalye2018synthesizingrobustadversarialexamples}.
For instance, in image classification,
researchers have identified adversarial images that can mislead
vision models, such as altering a stop sign to resemble a speed limit sign~\citep{eykholt2018robustphysicalworldattacksdeep}.
To address this issue, robustness specifications have been introduced---requiring
that, despite noise, the image should still be classified correctly~\citep{ai2,deepg,mirman2020robustnesscertificationgenerativemodels}.

However, unlike vision tasks, where some notions of correctness exist~\citep{athalye2018synthesizingrobustadversarialexamples,YangLM023},
determining correctness in system applications is more challenging. Typically, manually-designed specifications are hard to design, error-prone, and limited to some corner-cases.
Consider the example of a video bitrate controller:
given the current network conditions and buffered video frames,
it is difficult to manually label bitrate choices as ``incorrect''.
Prior work~\citep{verifying_learning_augmented_systems} has introduced specifications based on
extreme cases, excluding several desirable trustworthy behaviors as we observe in our experiments.
Furthermore, the authors of NN4SysBench~\citep{nn4sysbench}, a benchmark suite for NNs in systems,
acknowledge that their specifications are inherently subjective. 
Our aim is to mitigate these drawbacks of manually-designed specifications with automated specification generation for NNs in computer systems. We desire the specifications to be expressive, containing multiple aspects of correct/trustworthy behaviors in the given applications. Thus, we investigate:  
\noindent
\begin{center}
    \textbf{How can we automatically generate specifications encoding several trustworthy behaviors for NNs in computer systems?}
\end{center}

\textbf{Main idea}.
We observe that a unique characteristic of computer systems is the presence of
traditionally-used rule-based algorithms and heuristics---referred to as \emph{references}.
Although these references may not match the performance of neural
networks, they are considered reliable, having been crafted according to domain
experts' understanding of correct behavior and rigorously tested in production environments.
This forms an analogy: references in systems
serve a similar role as, for example, human perception does in image classification.
By collecting the outputs of these references across various inputs and
environments, we can define specifications for their neural network
counterparts.

Implementing this idea presents three key challenges.
First, multiple traditional implementations often exist for the same application;
for instance,
model predictive control~\cite{mpc}
and buffer-based algorithms~\cite{bba} are both widely used adaptive bitrate algorithms.
These implementations may exhibit divergent behaviors on the same input,
making it unclear how to consolidate their feedback into specifications for NNs.
Second, the set of specifications derived from references is not unique,
and some specifications are more useful than others.
Thus, we need to design criteria to identify \emph{high-quality} specifications.
Finally, synthesizing specifications is generally expensive and error-prone~\citep{wen2024enchantingprogramspecificationsynthesis,max_spec_synthesis,spec_inference}.
We need an efficient algorithm capable of generating
specifications with the desired traits.

\textbf{Our approach}.
We propose an automated specification generation method
for system domains where references are available.
Our specifications are based on pre and postconditions consisting of constraints
on the neural network inputs and outputs respectively~\cite{hoare:69}.
We specify that if the network input satisfies the
precondition, its output must satisfy the postcondition.
The postconditions are constructed by combining the behaviors of multiple references.

We formalize the desired traits that specifications should satisfy and pose their
generation as an optimization problem (\S~\ref{sec:formal}).
We design an effective algorithm, \emph{\tool} to
solve the optimization problem using clustering.
Our algorithm assumes access to offline observations (i.e., we do not use the source code or make custom queries to get observations of the behaviors of references),
therefore, our algorithm can handle references that have complicated
implementations, whose source code is unavailable or inference is either impossible or expensive.
We generate expressive and
useful specifications for NNs
for two challenging and practically important applications: adaptive bit rate streaming~\cite{pensieve} and congestion control~\cite{aurora}.

\textbf{Solution scope.}
As the first step towards automatically generated specifications for NNs in computer systems,
the specifications learned from references
serve as guidance rather than strict requirements.
In particular, some behaviors in reference
implementations should be emulated by NNs, while others should be
avoided, allowing NNs to potentially exceed the performance of conventional
approaches---achieving this balance is essential. 
In addition,
behavioral expectations are not absolute given the dynamically-varying environments in which computer systems are operated;
and similarly specifications are \emph{likely} encodings of the notion of trustworthiness in the given domain, as they are generated from observations of the references operated in these environments.
Nonetheless, the specifications are useful for testing and verification (\S\ref{sec:results}) and can serve as concise,
interpretable descriptions of desirable behaviors in the chosen domain. The specifications can also be used to compare different NNs; a network that satisfies the specification while achieving high performance is more aligned with developer expectations than one that does not.
%

\textbf{Contributions}. Our main contributions are: 
\begin{enumerate}
    \item We formalize generating expressive and useful specifications
        from references as an optimization problem. 

    \item We develop an automated specification generation algorithm, called \tool{}\footnote{\textbf{Spec}ifications from \textbf{T}rustworthy
        \textbf{R}eference \textbf{A}lgorithms} using our formalism, that
        can handle complicated and closed-sourced references. \tool{} leverages clustering to identify common behaviors across observations from the references. Code is available at \url{https://github.com/uiuc-focal-lab/spectra}.

    \item We create specifications for NNs in Adaptive Bitrate and Congestion
        Control applications. We empirically demonstrate (\S\ref{sec:results}) the high quality of our specifications. We also use \tool{}'s specifications to identify previously unknown vulnerabilities of the tested NNs.
\end{enumerate}

\tool is the first step towards formally verified NNs in computer systems. As \tool{}'s specifications are generated from references that are deemed trustworthy by domain experts, checking and enforcing NNs to adhere to the specifications can enhance the reliability of these models and encourage their practical deployment.

%% file: background.tex
\section{Background}
We demonstrate specifications for $2$ applications --- Adaptive Bit Rate (ABR) video streaming~\citep{abr_suvey} and Network Congestion Control (CC)~\citep{cc_survey}.

\textbf{Adaptive Bitrate}.
Adaptive Bit Rate (ABR) algorithms are used to optimize the bit rate for streaming videos from servers to clients such that the Quality of Experience, QoE~\citep{qoe} for the users is maximized. Quality of experience is typically determined by the bit rate of video chunks (higher is better), and the startup time, rebuffering time, and bitrate variations (lower is better). An ABR algorithm observes the video streaming system's state consisting of buffer size and video chunk download time, observed throughput, size of the next video chunk at all possible bit rates, etc. to determine the bit rate at which the next video chunk should be fetched. Pensieve~\cite{pensieve} is a popular neural network (NN) used for ABR. It is trained using reinforcement learning (RL) to maximize the QoE. We provide Pensieve's architectural details in Appendix~\ref{app:arch}.

\textbf{Congestion control}. Congestion control is a regulatory process that determines the packet sending rate across a given network at any time, to maximize the network throughput (packets sent over the network) and minimize latency and packet loss~\citep{aurora}. Congestion control algorithms use latency gradient~\citep{latgrad} and latency ratio~\citep{latratio} as input features and output the change in sending rate for the next time step. Aurora~\cite{aurora} is a popular NN-based solution for congestion control, trained with RL (architectural details in Appendix~\ref{app:trainingaurora}). The reward here is a combination of the throughput, latency, and packet loss observed in the system. 

%% file: technical.tex
\section{Formalizing specifications from reference algorithms}\label{sec:formal}
While neural networks attempt to maximize aggregate performance, we develop a set of specifications $\setspecs$ for individual inferences to satisfy for greater trust. 
Let $\inputs\subseteq\mathbb{R}^\numinputs$ be the sets of all possible $m$-dimensional ($m > 0$) inputs to the neural network for which we want to generate $\setspecs$. Let $\outputs = [1,\dots,\numoutputs]$ be the (discrete) set of all possible outputs of the neural network, where $1<\numoutputs < \infty$. 
Such instances with finite output sets are fairly common, e.g., in neural-network classifiers~\citep{classification1,mnist} and RL agents over finite action spaces such as Pensieve~\citep{pensieve}. If that is not the case, we discretize $\outputs$ when possible. For example, the output of the Aurora congestion control model~\citep{aurora} is continuous-valued and we generate specifications for it. We provide details of the output discretization for Aurora's specifications in \S\ref{sec:expt_setup}.
Each specification $\spec\in\setspecs$ describes the desirable behavior over a set of possible inputs called a \emph{precondition}, denoted by the Boolean predicate $\pre_\spec$ that evaluates to true for inputs in the precondition. $\spec$ mandates that for all inputs $x\in\inputs$ such that $\pre_\spec(x)$, the output $y\in\outputs$ should follow a \emph{postcondition}, denoted by the Boolean predicate $\post_\spec$, i.e., $\spec(x,y) \triangleq \pre_\spec(x)\implies\post_\spec(y)$. The specifications in $\setspecs$ are considered to be in conjunction, i.e., the overall desirable behavior is $\bigwedge_{\spec\in\setspecs}\spec$. 

We leverage the behavior of $q$ ($>0$) traditional algorithms, aka references $\abrs_1,\dots\abrs_\numabrs$ to determine $\setspecs$. We use multiple references to avoid $\setspecs$ that overfit the behavior of one reference and contain its suboptimal behaviors. Specifically, we consider inputs $x\in\inputs$ for which the combined set of outputs from all the references $\bigcup_{j\in[\numabrs]} \abrs_j(x)$ is non-trivial, i.e., excludes some possible outputs from $\outputs$. For other inputs, the references are not selective and do not provide useful information about the desirable behaviors in the target domain. Hence, they are not used to form the specifications. We use $\commonbeh$, the mapping between such $x$ and their corresponding outputs from references $\bigcup_{j\in[\numabrs]}\abrs_j(x)$, as the \emph{interesting behaviors} of references (Definition~\ref{def:nontrivialbeh}), which are used for generating specifications $\setspecs$. We hypothesize and empirically show that if a neural network's output matches that of any reference for inputs in the interesting behaviors, then it will be more trustworthy.

\begin{definition}
    \label{def:nontrivialbeh}
    (Interesting behaviors of references). Interesting behaviors $\commonbeh$ contain inputs $x\in\inputs$ for which all references $\abrs_1,\dots,\abrs_\numabrs$ collectively lead to a non-trivial output set $y\subset\outputs$, which is a strict subset of all possible outputs.
    $$\commonbeh \triangleq\left\{(x, y)\mid x\in\inputs\wedge y = \bigcup_{j\in[\numabrs]} \abrs_j(x)\wedge y\subset\outputs\right\}$$ 
\end{definition} 
We use $\commonbeh_x$ to denote the inputs $x$ in the interesting behaviors $\commonbeh$. A lookup table mapping $\commonbeh_x$ to the corresponding outputs from the references is an exact set of specifications. However, it is not amenable to downstream applications such as verification of the neural networks as it can potentially consist of uncountably many entries. The lookup table may not have finite representations in a general case. Therefore, we combine several $x\in\commonbeh_x$ into a single concise representation $\pre_\spec$ via overapproximation in each of our specifications $\spec$ and map it to $\post_\spec$ which captures the permissible output of $x$. 
The downside of simple and concise representations is that they can potentially introduce errors by restricting the outputs of the additional inputs $\hat{x}\notin\commonbeh_x$ that satisfy $\pre_\spec$ to $\post_\spec \subset \outputs$. We attempt to minimize such overapproximation errors when developing our specifications. 

Typically, concise representations consist of polyhedral constraints on permissible inputs, each of which is mapped to a common set of outputs~\citep{vnncomp23}. The simplest and widely used polyhedral representation consists of interval constraints. It leads to easily interpretable specifications. Hence, our specifications are defined using intervals that overapproximate interesting behaviors. 

\textbf{Preconditions}. The precondition of each specification consists of intervals over each dimension of the input space. Their canonical form is $\pre_\spec(x) = \forall i\in\{1,\dots,\numinputs\}\ldotp x_i\in[l_{i,\spec},r_{i,\spec}]$ ($x_i$ = $i$-th element of vector $x$) with $l_{i,\spec} \leq r_{i,\spec}$. Next, we list desirable properties of preconditions. 

\begin{itemize}
    \item \textbf{High representation}. We want each specification $\spec\in\setspecs$ to capture several interesting behaviors to be meaningful and important for $\setspecs$. With more interesting behaviors in each $\spec$, we will need fewer specifications in $\setspecs$, making it more interpretable. For this we define the \textit{representation} of a specification $\spec$ as: 
    \begin{equation}
    \label{eq:representation}
    Rep(\spec) \triangleq \frac{|\{x\mid x\in\commonbeh_x\wedge\pre_\spec(x)\}|}{|\commonbeh_x|}
\end{equation}
    We require each specification's precondition to contain at least $\threshrep\in(0,1]$ fraction of $\commonbeh_x$, i.e., $Rep(\spec) \geq\threshrep$, where $\threshrep$ is a user-defined threshold. 
    \item \textbf{High coverage}. We want all specifications in $\setspecs$ to collectively cover a large fraction of $\commonbeh_x$. This increases the interesting behavior information conveyed by $\setspecs$. We define the \textit{coverage} of $\setspecs$~\eqref{eq:coverage} as the fraction of $\commonbeh_x$ captured by any specification's precondition in $\setspecs$.\begin{equation}
    \label{eq:coverage}
    Cov(\setspecs) \triangleq \frac{\mid\bigcup_{S\in\setspecs}\{x\mid x\in\commonbeh_x\wedge\pre_S(x)\})\mid}{\mid\commonbeh_x\mid} 
\end{equation} We desire a coverage more than a user-specified threshold $\threshcov\in(0,1]$, i.e., $Cov(\setspecs) \geq\threshcov$. 
    \item \textbf{Low volume}. To reduce overapproximation error due to interval-based preconditions, we want to reduce the number of inputs allowed by any specification $\spec\in\setspecs$ while maintaining high coverage and representation scores. This can be done by minimizing the volume of $\setspecs$ which is the sum of the volume of each $\pre_\spec$~\eqref{eq:vol}. The parts of $\inputs$ accepted by $\pre_\spec$ are hyperrectangles denoted by the intervals $[l_{i,\spec}, r_{i,\spec}], \forall i\in\{1,\dots,\numinputs\}$. Hence their volumes are defined as products of their ranges along each dimension. \begin{equation}
    \label{eq:vol}
    Vol(\setspecs) \triangleq \sum_{\spec\in\setspecs} \prod_{i\in\{1,\dots,\numinputs\}} (r_{i,\spec} - l_{i,\spec})
\end{equation}
\end{itemize}

\textbf{Postconditions}. Let $\pre_\spec(\commonbeh_x)\triangleq \{x\in\commonbeh_x \mid\pre_\spec(x)\}$ denote the interesting behavior inputs that satisfy $\pre_\spec$. We define the postcondition $\post_\spec$ for a given precondition $\pre_\spec$ as $\post_\spec(y) \triangleq y\in\bigcup_{x\in\pre_\spec(\commonbeh_x)}\bigcup_{j\in[\numabrs]}\abrs_j(x)$, i.e., $\post_\spec$ accepts all the outputs of all the references for the interesting behaviors captured by $\pre_\spec$. Let $\card_\spec = |\{y\mid y\in\outputs\wedge\post_\spec(y)\}|$ denote the number of allowed outputs by $\spec$. Ideally, we should map each input $x\in\pre_\spec(\commonbeh_x)$ only to its permissible output from the references $\bigcup_{j\in[\numabrs]}\abrs_j(x)$. However, this can be difficult to satisfy with the coverage and representation constraints on precondition. Hence, we relax this requirement to overapproximate the allowed outputs of $x$ with outputs permissible for other elements of $\pre_\spec(\commonbeh_x)$. To minimize the overapproximation error, we do not allow more than a specific number of outputs to be accepted by $\post_\spec$, given by a threshold $\threshtight\in\{1,\dots,|\outputs-1|\}$, i.e., $\card_\spec\leq\threshtight$. 


\textbf{Optimization problem}. For high-quality specifications, we generate specification sets $\setspecs$ satisfying a minimum coverage threshold $\threshcov$, with each specification having a representation score higher than a threshold $\threshrep$ and having a maximum of $\threshtight$ outputs in the postcondition. With these constraints, we want to minimize the volume of $\setspecs$. The optimal specifications set $\setspecs^*$ is the solution to the optimization problem in Equation~\eqref{eq:overalloptim}. We allow the thresholds $\threshcov,\threshrep,\threshtight$ to be user-defined to generalize to varying domain-specific requirements for coverage, representation, and maximum number of outputs in postconditions.

\begin{align}
    \label{eq:overalloptim}
    &\setspecs^* = argmin_{\setspecs} Vol(\setspecs) \\ 
    \notag &s.t. \quad Cov(\setspecs) \geq \threshcov,\quad \forall\spec\in\setspecs\ldotp Rep(\spec) \geq \threshrep,\\ \notag &\forall\spec\in\setspecs\ldotp \card_\spec \leq \threshtight
\end{align}

%% file: practical_technical.tex
\section{\tool --- Specifications from Reference Algorithms}\label{sec:implementation}
In this section, building on our formalism from Section~\ref{sec:formal}, we describe our algorithm \tool, for automatically generating high-quality specifications from reference algorithms.

\subsection{Practical optimization problem}
The optimization problem~\eqref{eq:overalloptim} is hard to solve in general settings, as identifying all the interesting behaviors for any given references is not trivial. The references can be complex functions and may lack closed-form expression. To identify the interesting behaviors across multiple references, we need to encode them in languages such as SMT-Lib~\cite{BarFT-SMTLIB} to use specialized solvers, such as Z3~\cite{z3}, which requires extensive manual efforts and may not be feasible for complicated references based on thousands of lines of intricate low-level code (e.g., for congestion control). Thus, for general applicability and ease of usage of our framework, we only assume access to the references through availability of some of their observations, from which we identify the interesting behaviors. Note that we do not assume the ability to control the observations by not assuming query-access to the references unlike prior specification generation works~\citep{precond_syn,synthesizing_contracts}. This causes us to use the observations available from the references in production settings, while reducing the costs of running them in sandboxed environments. Moreover, the observations contain the behaviors of the references seen during practical deployment, which supports our quest for specifications encoding practical reliability.
However, the assumption of a static, given dataset of observations imposes several constraints and \tool{} applies the following adaptations to the optimization problem in~\eqref{eq:overalloptim} to solve it in this setting. 


\textbf{Interesting input \emph{regions}}. 
A drawback of our assuming a static, given dataset of observations is that we cannot ensure the availability of the outputs of all references for any specific input $x\in\inputs$, to identify interesting behaviors. However, we may have observations from references for `close-by' inputs, which can indicate the behaviors of the references in a \emph{local} input region. Let $X$ be such an input region containing $x_1,\dots,x_i,\ldots  \in\inputs$ observations. The reference $\abrs_j$'s output $Y_{X,j}\subseteq\outputs$ for $X$ is the set of outputs of $\abrs_j$ for any input in $X$ in the given observations, i.e., $Y_{X,j} \triangleq \abrs_j(X) = \bigcup_{x\in\inputs}\abrs_j(x)$. We treat a local region $X$ as a single entity for the following discussion.

\begin{definition}
(Interesting behavior regions)  Interesting behavior regions $\commonbehreg_X \triangleq \{\dots,X_i,\dots\}$ is a potentially infinite set of non-overlapping local regions $X_i$,  where each $X_i$ contains multiple observations from each reference and $Y_{X_i} = \bigcup_{j\in[\numabrs]}Y_{X_i,j}\subset\outputs$. $\commonbehreg_y$ denotes the set of outputs corresponding to $\commonbehreg_x$, i.e., $\commonbehreg_y = \{\dots,Y_{X_i},\dots\}$. 
\end{definition}
%

The preconditions of specifications, $\pre_\spec$ are still based on intervals and an input region $X$ is accepted by  $\pre_\spec$ when all points in $X$ satisfy the precondition $\pre_\spec(X) \iff \forall x\in X\ldotp\pre_\spec(x)$. An output $y$ satisfies the postcondition if it is included in the output of a $X \in \commonbehreg_X$ for which $\pre_{\spec}(X)$ is true, i.e., $\post_\spec(y) \triangleq (y\in\bigcup_{X\in\commonbehreg_X\wedge\pre_\spec(X)}Y_X$). Interesting behavior regions overapproximate ideal interesting behaviors introducing overapproximation errors as we map each $X\in\commonbehreg_X$ to outputs based on the observations in $X$, which may exclude some outputs for inputs of $X$ that are not observed. We require the interesting behavior regions to be generated with more observations to reduce the error. Note that such interesting behavior regions are similar to robustness regions around given inputs as defined and used in manually designed specifications in several prior works~\citep{Chakravarthy2022PropertyDrivenEO,dnn_specs}. However, the salient difference from the latter is that the interesting behavior regions are automatically determined using several observations of references.

\textbf{Relaxed metrics}. To incorporate the above notion of interesting behavior regions, $\commonbehreg_X$, we adapt our desirable properties of high representation~\eqref{eq:representation} and high coverage~\eqref{eq:coverage} metrics. 
Originally coverage denotes the fraction of interesting behaviors captured by a specifications set $\setspecs$. In this case, coverage modifies to the fraction of $\commonbehreg_X$ captured in $\setspecs$~\eqref{eq:relcoverage}. We desire a coverage higher than a given threshold in the relaxed problem $\widetilde{\threshcov}$, i.e., $\widetilde{Cov}(\setspecs) \geq \widetilde{\threshcov}$. 

\begin{equation}
    \label{eq:relcoverage}
    \widetilde{Cov}(\setspecs) \triangleq \frac{\mid\bigcup_{S\in\setspecs}\{X\mid X\in\commonbehreg_X\wedge\pre_\spec(X)\})\mid}{\mid\commonbehreg_X\mid}
    \end{equation}

Similarly, representation relaxes to the fraction of $\commonbehreg_X$ captured in a given specification~\eqref{eq:relrepresentation}. We want it to be more than a given threshold $\widetilde{\threshrep}$, i.e., $\forall\spec\in\setspecs\ldotp\widetilde{Rep}(\spec) \geq \widetilde{\threshrep}$. 
\begin{equation}
        \label{eq:relrepresentation}
        \widetilde{Rep}(\spec) \triangleq \frac{\mid\{X\mid X\in\commonbehreg_X\wedge\pre_\spec(X)\})\mid}{\mid\commonbehreg_X\mid}
    \end{equation}

The final practical optimization problem~\eqref{eq:practicaloptim} thus minimizes the volume of the specifications set, while covering at least $\widetilde{\threshcov}$ fraction of interesting behavior regions overall, with each specification covering at least $\widetilde{\threshrep}$ fraction of interesting behavior regions and allowing less than $\threshtight$ outputs.

\begin{align}
    \label{eq:practicaloptim}
    &\widetilde{\setspecs^*} = argmin_{\setspecs} Vol(\setspecs) \\ 
    \notag &s.t. \quad \widetilde{Cov}(\setspecs) \geq \widetilde{\threshcov},\quad \forall\spec\in\setspecs\ldotp \widetilde{Rep}(\spec) \geq \widetilde{\threshrep},\\  &\notag\forall\spec\in\setspecs\ldotp \card_\spec \leq \threshtight
\end{align}

\subsection{Implementation}
\begin{figure*}[tb]
    \centering
    \includegraphics[width=0.9\linewidth]{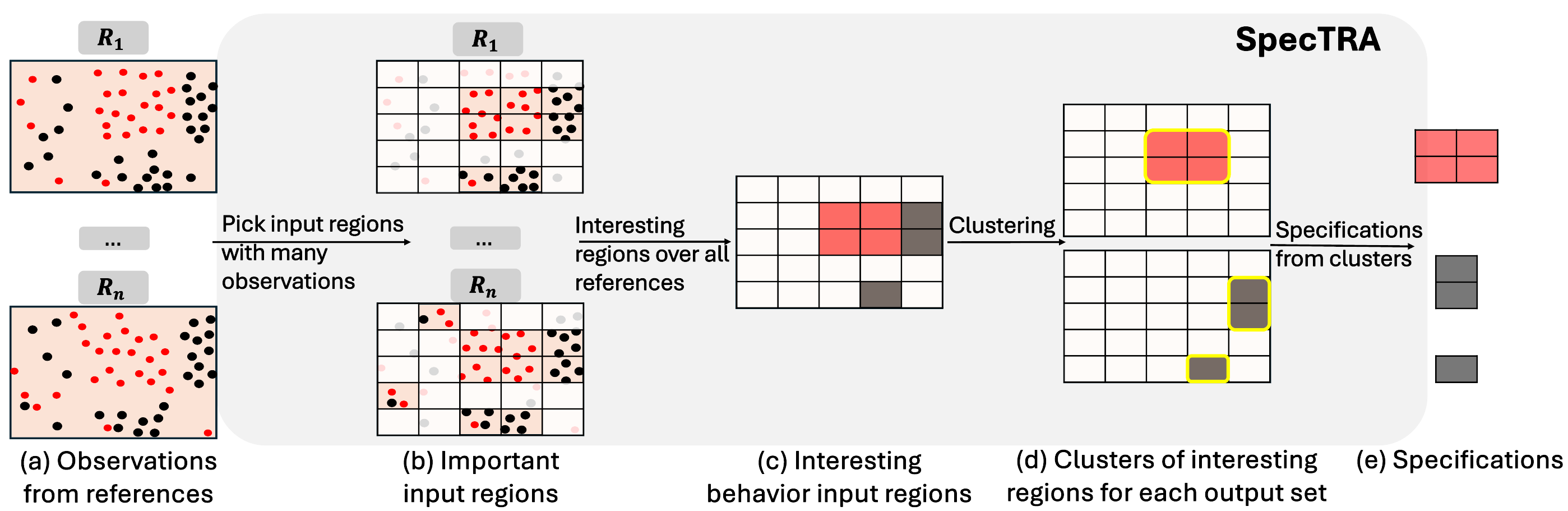}
    \caption{(Overview of \tool{}) Observations from all the references (a) are given as input to our specification generation algorithm, \tool{}. For illustration purposes, a 2-D feature space is considered and each input is classified in either the red or black class by the corresponding reference algorithm. \tool{} takes the observations and from them develops a set of likely specifications (e). To generate the specifications, \tool{} first partitions the input space into input regions and from them identifies the regions which have several observations for each reference separately (b) to form the important input regions for the reference. These important regions are combined into interesting behavior input regions (c) according to Definition~\ref{def:nontrivialbeh}. Clusters are identified in the interesting behavior regions with the same output set according to density (d). These clusters form the specifications (e) that map the inputs in each cluster to the corresponding output set of the cluster.}
    \label{fig:overview}
\end{figure*}

Figure~\ref{fig:overview} presents a high-level overview of \tool{}'s working. 
Algorithm~\ref{alg:main} gives the pseudocode of \tool{}'s algorithm to solve~\eqref{eq:practicaloptim}. Let $\data_{1},\dots\data_{\numabrs}$ be the set of observations from the references $\abrs_1,\dots,\abrs_\numabrs$ respectively.

\textbf{Identifying interesting behavior regions}. To identify the interesting behavior regions across references, Algorithm~\ref{alg:main} first identifies non-overlapping local input regions, which could potentially be interesting, from $\inputs$. We form the input regions by partitioning $\inputs$ along each dimension into a fixed number $\parts>0$ of equally-sized parts (line~\ref{algline:partition}). Input regions may be identified with other partitioning methods too, however, \tool{} is agnostic to them. We selected this partitioning method for simplicity. Not all input regions thus obtained can be worth considering, as some may have very few or $0$ observations of some references. Therefore, we identify the important input regions having at least a certain fraction of the available observations for every reference (line~\ref{algline:impregions}) to reduce the overapproximation error due to considering input regions. We then obtain the interesting behavior regions from the important regions common across all the references by applying the condition for interesting behaviors (Definition~\ref{def:nontrivialbeh}). We thus obtain the set of input regions $\commonbehreg_X$ and their corresponding outputs $\commonbehreg_y$ from $\data_{1},\dots,\data_{\numabrs}$ as interesting behavior regions (line~\ref{algline:interestingregions}).
\begin{algorithm}[!tb]
   \caption{\tool{} Algorithm}
   \label{alg:main}
\begin{algorithmic}[1]
   \STATE {\bfseries Input:} $\inputs$, $\outputs$, $\data_{1},\dots\data_{\numabrs}$, $\widetilde{\threshcov}$, $\widetilde{\threshrep}$, $\threshtight$, $\parts$
   \STATE {\bfseries Output:} $\setspecs$
   \STATE $\setspecs\gets\emptyset$
   
   \COMMENT{ \textcolor{blue}{Identifying interesting behavior regions}}
    \STATE input\_regions $\gets$ \texttt{Partition}($\inputs$, $\parts$)\label{algline:partition}
    \STATE $\mathcal{I}$ $\gets$ \texttt{Important}(input\_regions, $\data_{1},\dots,\data_{\numabrs}$)\label{algline:impregions}
    \STATE $\commonbehreg_X$, $\commonbehreg_y$ $\gets$ \texttt{Interesting}($\mathcal{I}$)\label{algline:interestingregions}
    
   \COMMENT{\textcolor{blue}{Combining interesting behavior regions}}
    \FOR{$i\in[1,\dots,\threshtight]$}
        \FOR{each subset \(\outsubset \subset \outputs\) with \(|\outsubset| = i\)}
           \STATE $\commonbehreg_{\outsubset,x}$ $\gets$ $\{X\mid X\in\commonbehreg_X, Y_X\subseteq \outsubset\}$\label{algline:commonbehx}
           \STATE $\commonbehreg_{\outsubset,y}\gets\{Y_X\mid X\in\commonbehreg_X\}$\label{algline:commonbehy}
           \STATE $\mathcal{L} \gets$ \texttt{Cluster}($\commonbehreg_{\outsubset,x}$,$\widetilde{\threshrep}$)\label{algline:cluster}
           \STATE $\spec_1,\dots,\spec_l \gets$ \texttt{Cluster2Spec}($\mathcal{L}, \commonbehreg_{\outsubset,x},\commonbehreg_{\outsubset,y}$)\label{algline:makespecs}
           \STATE $\setspecs.extend(\spec_1,\dots,\spec_l)$\label{algline:addspecs}
           \IF{$\widetilde{Cov}(\setspecs)\geq\widetilde{\threshcov}$}\label{algline:cov}
                \STATE {\bfseries return} $\setspecs$
           \ENDIF\label{algline:covend}
        \ENDFOR 
    \ENDFOR
   \STATE {\bfseries return} $\setspecs$
\end{algorithmic}
\end{algorithm}

\textbf{Combining interesting behavior regions into specifications}. 
The identified interesting behavior regions are used to solve the optimization problem in~\eqref{eq:practicaloptim}. Firstly, \tool{} considers the constraint on the maximum number of outputs accepted by the postconditions, i.e., $\forall\spec\in\setspecs\ldotp \card_\spec \leq \threshtight$. To satisfy the constraint, \tool{} forms individual specifications only over $X$ ($\in\commonbehreg_X$) which combine to yield a postcondition accepting less than $\threshtight$ outputs. To do so, \tool{} enumerates all subsets $\outsubset$ of $\outputs$, such that $|\outsubset|\in[1,\dots,\threshtight]$. For each $\outsubset$, it filters out the interesting behavior regions having output as a subset of $\outsubset$ (lines~\ref{algline:commonbehx}-\ref{algline:commonbehy}). The problem thus reduces to generating specifications for $\commonbehreg_{\outsubset,x}, \commonbehreg_{\outsubset,y}$ without any constraints on the postconditions. 

The precondition of each specification that allows any subset of $\commonbehreg_{\outsubset,x}$ should be the tightest bounding hypercube the subset, to minimize the volume of the specification. Moreover, as all input regions have the same volume, minimizing the volume of specification sets reduces to minimizing the number of input regions allowed by the preconditions of the specifications, while satisfying the representation and coverage constraints. The optimization problem becomes minimizing the number of input regions in the preconditions (volume) while retaining a minimum number of interesting behavior input regions in every specification (representation) and over all the specifications (coverage). Exactly solving this optimization problem is hard, as (i) the search space is large with several observations, high number of parts $p$ along each input dimension, high number of input dimensions, (ii) there are no obvious structures (e.g., decomposability or differentiability) that we can exploit in our setting where we cannot query or access the reference implementations. Hence, we develop an approximate solution by identifying clusters in $\commonbehreg_{\outsubset,x}$ according to the proximity of constituent regions. To minimize the volume, we apply density-based clustering methods such as the DBSCAN algorithm~\citep{dbscan} as they find dense clusters and are resistant to outliers. The general \texttt{Cluster}(.) function in line~\ref{algline:cluster} uses the selected clustering algorithm to returns $l$ clusters from $\commonbehreg_{\outsubset,x}$, denoted as cluster labels $\mathcal{L}\in[1,\dots,l]^{|\commonbehreg_{\outsubset,x}|}$ for each element in $\commonbehreg_{\outsubset,x}$. We encode the representation constraint into the clustering process, requiring each cluster to have at least $\widetilde{\threshrep}$ samples. Each cluster thus identified can be combined to form a specification (line~\ref{algline:makespecs}). Specifically, each $\spec_i, i\in[1,\dots,l]$ has $\pre_{\spec_i}$ as the tightest hyperrectangle bounding $\mathcal{R} \triangleq\{X\mid X\in\commonbehreg_{\outsubset,x}, \mathcal{L}(X) = i\}$, where $\mathcal{L}(X)$ denotes the cluster label for input region $X$. $\post_{\spec_i} = \bigcup_{X\in\mathcal{R}}Y_X$, i.e., the post condition allows for the outputs corresponding to any element of $\commonbehreg_{\outsubset,x}$ in the $i^{th}$ cluster. For the coverage constraint, we adopt a best-effort approach (lines~\ref{algline:cov}-\ref{algline:covend}), wherein we keep generating specifications till either $\widetilde{\threshcov}$ is achieved or all options for $\outsubset$ are exhausted. 

%% file: evaluation.tex
\section{Experiments}\label{sec:results}

In this section, we study the \emph{quality} and \emph{utility} of \tool{}'s specifications. We illustrate \tool{}'s specifications in two applications having reference algorithms and their neural counterparts --- Adaptive Bit Rate (ABR) algorithms in video streaming and Congestion Control (CC) algorithms. Pensieve~\citep{pensieve} is a popular neural network (NN) based RL-agent for ABR that decides the bit rate for the next video chunk. Pensieve needs to choose from $6$ possible bit rates for the next video chunk --- $300, 750, 1200, 1850, 2850, 4300$ kbps. While ABR has several traditional (reference) algorithms, there are two salient ones, that we use as references for generating specifications for Pensieve --- the Buffer-based (BB) algorithm~\citep{bba} and the Model Predictive Control (MPC) algorithm~\citep{mpc}. Aurora~\citep{aurora} is a popular NN-based RL-agent for CC that proposes real-valued changes to the rate of sending packets over a network to reduce congestion. The CC references that we consider for generating specifications for Aurora are the BBR~\citep{bbr} and Cubic~\citep{cubic} algorithms. 
Note that our framework is general to handle more than two references. However, we expect to get fewer interesting behavior regions with additional references, which may deteriorate the quality of the specifications. Hence, we choose to limit to two references for each application.
As \tool{} uses several thresholds and parameters to generate specifications, we show an ablation study (Appendix~\ref{sec:ablation}) to know their effects on the specifications' quality and generate specifications with the best settings. 

\subsection{Experimental setup}\label{sec:expt_setup}
We conducted our experiments on a 12th Gen 20-core Intel i9 processor. 
We collect the observations from references in the training environments of the NN, to generate relevant specifications. For ABR, as we have the implementations of the references readily available, we run them in the training environment of the target model Pensieve, with its training traces that govern the network characteristics for video streaming at each time step. We use the publicly available training dataset of Pensieve consisting of $128$ traces, each having 100s of time steps. We test the specifications on Pensieve's test set that consists of $143$ traces, also having 100s of time steps each.
For CC, however, the reference algorithms are embedded in the operating system kernels, making them less amenable to run in the target model Aurora's training environment. Hence, we obtain observations for them from their execution logs in the Pantheon project~\citep{pantheon}. We retrain the Aurora models using some of the corresponding network traces from Pantheon so as to align the reference observations with the training environment of the models. We use $75\%$ traces for training and remaining for testing the Aurora models. We describe the details of the retraining and mention the specific traces used in training and testing in Appendix~\ref{app:trainingaurora}. We experiment with both the retrained and original Aurora models.
We generate specifications for both applications with the observations from references on the training traces and use the observations from the testing traces to evaluate the specifications. 
\begin{table*}[tb]
    \centering
    \small{\begin{tabular}{@{}l|lp{2cm}>{\raggedleft\arraybackslash}p{2cm}>{\raggedleft\arraybackslash}p{2cm}>{\raggedleft\arraybackslash}p{1.5cm}>{\raggedleft\arraybackslash}p{1.5cm}@{}}
    \toprule
        Application & Reference & Observation type & Support (\tool{}) & Confidence (\tool{}) & Support (prior) & Confidence (prior) \\
    \midrule
        \multirow{4}{*}{ABR} & \multirow{2}{*}{BB} & Training & $0.95$ & $1.0$ & $0.03$ & $1.0$ \\
        & & Test & $0.96$ & $1.0$ & $0.01$ & $0.99$ \\
    \cline{2-7}
         & \multirow{2}{*}{MPC} & Training & $0.64$ & $0.87$ & $0.09$ & $0.87$ \\
        &  & Test & $0.59$ & $0.87$ & $0.07$ & $0.84$\\
    \cline{2-7}
         & \multirow{2}{*}{Pensieve (small)} & Training & $0.17$ & $0.99$ & $0.27$ & $0.86$ \\
        &  & Test & $0.13$ & $0.99$ & $0.25$ & $0.85$\\
    \cline{2-7}
         & \multirow{2}{*}{Pensieve (mid)} & Training & $0.31$ & $0.96$ & $0.14$ & $0.93$ \\
        &  & Test & $0.26$ & $0.97$ & $0.13$ & $0.91$\\
    \cline{2-7}
         & \multirow{2}{*}{Pensieve (big)} & Training & $0.26$ & $0.97$ & $0.22$ & $0.93$ \\
        &  & Test & $0.21$ & $0.98$ & $0.18$ & $0.92$\\
    \midrule
        \multirow{4}{*}{CC} & \multirow{2}{*}{BBR} & Training & $0.89$ & $0.75$ & $0.01$ & $0.43$ \\
        &  & Test & $0.74$ & $0.81$ & $0.01$ & $0.38$ \\
    \cline{2-7}
        & \multirow{2}{*}{Cubic} & Training & $0.97$ & $0.82$ & $0.001$ & $0.29$ \\
        &  & Test & $0.79$ & $0.79$ & $0.07$ & $0.42$\\
    \cline{2-7}
         & \multirow{2}{*}{Aurora (small)} & Training & $1.0$ & $1.0$ & $0.01$ & $1.0$ \\
        &  & Test & $1.0$ & $1.0$ & $0.01$ & $1.0$\\
    \cline{2-7}
         & \multirow{2}{*}{Aurora (mid)} & Training & $0.98$ & $1.0$ & $0.17$ & $0.03$ \\
        &  & Test & $0.96$ & $1.0$ & $0.34$ & $0.01$\\
    \cline{2-7}
         & \multirow{2}{*}{Original Aurora} & Training & $1.0$ & $1.0$ & $0.01$ & $1.0$ \\
        &  & Test & $0.99$ & $1.0$ & $0.05$ & $0.05$\\
    \bottomrule
    \end{tabular}}
    \caption{Support and confidence of \tool{}'s specifications and that of the specifications in~\citep{verifying_learning_augmented_systems} on the training and testing observations of the references and NNs for both the ABR and Congestion Control (CC) applications}
    \label{tab:supportconf}
\end{table*}

As \tool{} assumes a finite discrete set of outputs, which is not the case for Aurora (output is real-valued change of packet sending rate), we discretize the output using the sign function, which gives the sign of the change resulting in $3$ possible outcomes: `+', `-', and `0'.
\tool{} uses the DBSCAN~\citep{dbscan} clustering algorithm to solve the optimization problem in~\eqref{eq:practicaloptim}. We detail the settings of DBSCAN in Appendix~\ref{app:dbscan}.
\tool{} develops the specifications on a subset of input features of the target NN for which we empirically observe the best quality of specifications. For ABR, \tool{} uses the current buffer size and the download times observed in the last $3$ time steps. For CC, \tool{} uses the history of the latency gradient, latency ratio, and sending ratio features used by Aurora, over previous $4$ time steps. We have selected these specific features for the two applications following those in the manually-designed specifications in \citet{verifying_learning_augmented_systems} and selected the history of the features in specifications with an ablation study in Appendix~\ref{sec:ablation}. We keep the coverage threshold, $\threshcov$ to be $1$, so as to get specification sets with the highest possible coverage. \tool{} generates specifications for either application in less than 30 seconds.

\subsection{Quality of Specifications}
\subsubsection{Quantitative analysis}
To evaluate the quality of the specifications, we check them against the observations from the references over the training and testing environments for the neural models. We use the specification evaluation metrics of \emph{support} and \emph{confidence} inspired from prior specification mining work,~\citet{texada} and data mining literature~\citep{datamining}. We formally define these metrics next. 

We want the specifications to cover most of the observations $\data_{j}$ from each reference $\abrs_j$ to correctly describe their behavior. For this, we measure the fraction of observations from $\data_j$ that are accepted by the precondition $\pre_\spec$ of any specification $\spec$ in \tool{}'s generated specifications set $\setspecs$. This quantity is the support $\support(\setspecs, \abrs_j)$~\eqref{eq:support} of $\setspecs$ for $\abrs_j$.
\begin{equation}\label{eq:support}
    \support(\setspecs, \abrs_j) \triangleq \frac{|\{x|(x,\abrs_j(x))\in\data_{j}\wedge\bigcup\limits_{\spec\in\setspecs}\pre_\spec(x)\}|}{|\data_{j}|}
\end{equation}
Let $\data_{j,\setspecs}\triangleq\{x|(x,\abrs_j(x))\in\data_{j}\wedge\bigcup_{\spec\in\setspecs}\pre_\spec(x)\}$ denote the observations from reference $\abrs_j$ contributing to the support for $\setspecs$. 
Alongside support, we want the specifications to be correct on the observations. Thus, we check the fraction of instances in $\data_{j,\setspecs}$ where $\forall\spec\in\setspecs\ldotp \pre_\spec\implies\post_\spec$, i.e., postconditions of all specifications are satisfied whose preconditions hold for observations contributing to the support of $\setspecs$. We call this the confidence $\conf(\setspecs, \abrs_j)$ for $\setspecs$~\eqref{eq:confidence}. 
\begin{align}
    \label{eq:confidence}
    &\conf(\setspecs, \abrs_j)\\
    &\notag\triangleq\frac{|\{x|x\in\data_{j,\setspecs}\wedge {\forall\spec\in\setspecs}\ldotp(\pre_\spec(x)\implies\post_\spec(\abrs_j(x)))\}|}{|\data_{j,\setspecs}|}
\end{align}

We report the support and confidence for \tool{}'s specifications for both applications in Table~\ref{tab:supportconf}. 
We compare our specifications with those given by prior works~\citep{vnncomp23,verifying_learning_augmented_systems}. We find that the specifications in VNN-COMP 2023~\citep{vnncomp23} have 0 support over the observations for the references in both applications. Hence, we do not include them in our study. The specifications in~\citep{verifying_learning_augmented_systems} are temporal in nature and, therefore, not directly comparable. Hence, we use the negation of their specifications for bad system states for individual transitions and compare with our specifications. We present the support and confidence of the prior specifications in~\citet{verifying_learning_augmented_systems} in Table~\ref{tab:supportconf}. The total number of training and testing observations for the ABR algorithms are $77981$ and $27837$ respectively. We filter out the observations for CC to consist of those that fall within the  observation space on which the Aurora models are typically trained. The total number of training and testing observations for the CC references are $\sim130k$ and $\sim40k$ respectively, and those for the Aurora models are $\sim140k$ and $\sim200k$ respectively.


The specifications in the prior work have low support for the observations from the references and similar confidence as our specifications. These results indicate that our specifications correctly encode more of the trusted behaviors of the references over the relevant (training and testing) input distributions of the NNs than the existing specifications. The existing specifications show comparable support over the training and testing observations from Pensieve models but lower confidence than our specifications. 

\subsubsection{Qualitative analysis with case studies}
\textbf{ABR}. \tool{} generates a specification set with $30$ specifications for Pensieve models. We present some specifications from the generated set next, to demonstrate their quality and conformance with intuitively correct behavior. Note that all specifications are in conjunction, so they need to hold simultaneously for the satisfaction of the generated specifications set. We give the entire specification set for Pensieve in Appendix~\ref{app:specsets}. Note that the ranges of buffer size (BS --- duration of pre-retrieved video stored in the buffer) and video chunk download-time (DT) features are $[4, 60]$ seconds and $(0,\infty)$ respectively. Each video chunk is $4$ seconds long.

Specification~\ref{spec:abr}a shows the conditions for which the lowest $2$ bitrates --- 300 and 750 kbps are allowed by the postcondition. These conditions consist of low BS and $> 2$ seconds of DTs. Intuitively, the ABR algorithm should output low bit rates for such states of video streaming systems, as the buffer does not have enough video chunks to render and there is some delay in downloading new video chunks. In such scenarios, to prevent rebuffering, the video chunks should be fetched at lower bit rates. Manually-designed specifications, such as those in~\citet{verifying_learning_augmented_systems} capture only extreme behaviors, such as situations when the lowest bit rate must be output by the ABR algorithm. However, \tool{}'s specifications can encode intermediate behaviors, such as cases when the lowest $2$ bit rates can be permissible, as well. Specification~\ref{spec:abr}b shows cases where we specify that the ABR algorithm does not output the lowest bit rate. We specify that the buffer should contain more than $2$ video chunks, each of which is $4$ seconds long, and the DT should be a moderate value for the lowest bit rate to not get selected. Note that, this specification supplements the intuitive specification about avoiding the lowest bit rate in~\citet{verifying_learning_augmented_systems}. The prior work's specification forbids the lowest bit rate when BS is $> 4$ seconds and the DTs are $< 4$ seconds, whereas \tool{}'s specification disallows the lowest bit rate even when DTs can be $> 4$ seconds, with large enough buffer. The prior work does not specify the desirable behavior for $> 4$ seconds of DT with BS $> 4$ seconds. Moreover, the permissible ranges of the input features at different points in their history can vary in the preconditions of automatically generated specifications, unlike those in manually-designed, intuition-based specifications. Obtaining such fine-grained specifications is beyond the scope of manually-designed specifications but can be achieved using automated methods such as \tool{}.
\vspace{-0.5cm}
\begin{code}
\begin{center}
\begin{minipage}{0.14\textwidth}
\centering
\small{\begin{align*}
    &\textbf{\text{Precondition}}\\
    & BS\in[4.0, 5.0],\\
    & DT[-1]\in[2.8, 6.6],\\
    & DT[-2]\in[2.8, 6.6],\\
    & DT[-3]\in[5.4, 9.2]\\
    &\textbf{\text{Postcondition}}\\
    & BR\in\{300, 750\}
\end{align*}}
\subcaption{}
\end{minipage}
\hspace{0.13cm}
\begin{minipage}{0.24\textwidth}
\centering
\small{\begin{align*}
    &\textbf{\text{Precondition}}\\
    & BS\in[10.9, 12.3],\\
    & DT[-1]\in[4.1, 7.9],\\
    & DT[-2]\in[1.5, 6.6],\\
    & DT[-3]\in[1.5, 5.4]\\
    &\textbf{\text{Postcondition}}\\
    & BR\in\{750, 1200, 1850, 2850, 4300\}
\end{align*}}
\subcaption{}
\end{minipage}
\captionof{listin}{Conjunctive specifications for ABR. (BS: Buffer Size, DT[$-i$]: $i^{th}$ last download time, BR: Bit Rate)}
\label{spec:abr}
\end{center}
\end{code}
\vspace{-0.5cm}
\begin{code}
\begin{center}
\small{\begin{align*}
    &\textbf{\text{Precondition}}\\
    & LG[-1]\in[-1.0, 0.29], LG[-2]\in[-0.78, 0.07], \\
    & LG[-3]\in[-0.78, 0.29], LG[-4]\in[-1.0, 0.29]\\
    & LR[-1]\in[1.0, 1.88], LR[-2]\in[1.0, 1.88], \\
    & LR[-3]\in[1.0, 1.88], LR[-4]\in[1.0, 1.88]\\
    & SR[-1]\in[0.0, 17.18], SR[-2]\in[0.0, 17.18], \\
    & SR[-3]\in[0.0, 17.18], SR[-4]\in[0.0, 17.18]\\
    &\textbf{\text{Postcondition}}\\
    & \texttt{Change in Sending Rate}\in\{+, -\}
\end{align*}}
\captionof{listin}{For CC. (LG: latency gradient, LR: latency ratio, SR: sending ratio, x[$-i$]: $i^{th}$ last value of x)}
\label{spec:cc}
\end{center}
\end{code}
\textbf{CC}. \tool{} generates $1$ specification, shown in Specification~\ref{spec:cc}, for Aurora models. Interestingly, this specification's precondition contains the precondition of the specification allowing non-zero change of sending rate in its postcondition in~\citet{verifying_learning_augmented_systems}. The latter specification comprises of a very small fraction of \tool{}'s specification. For example, all latency gradients are specified to be within $[-0.01,0.01]$, latency ratios in $[1.0, 1.01]$, and sending ratios as only $1.0$. Thus, the corresponding prior specification is conservative, probably due to its manual design.

\subsection{Utility of Specifications}
\subsubsection{NN Verification}
Next, we explore a popular downstream application of NN specifications --- verifying trained NNs. We attempt to verify the NNs in each application for \tool{}'s specifications using the SOTA complete-verifier, $\alpha\beta$-CROWN~\citep{abcrown}. As the overall specification set is a conjunction of all elements in the specification set generated by \tool{}, we attempt to verify each specification in the set individually by encoding it in the VNN-Lib format~\citep{vnn-lib}. \tool{} generates a specification set containing $30$ specifications for ABR and $1$ specification for the CC setting. We set $\alpha\beta$-CROWN's timeout as 10 minutes. We use the more precise activation splitting for the Pensieve models and input-splitting for Aurora models. This is because $\alpha\beta$-CROWN does not support activation splitting for regression models currently, to the best of our knowledge. As our specifications encode only a subset of the inputs in the preconditions, we specify the other input features as their ranges as seen in the observations used to generate the specifications. Table~\ref{tab:nnverification} presents our findings for both applications. The \emph{Verified} instances occur when the specification is satisfied by the model, the \emph{Falsified} instances are when we can find a successful attack for the specification on the model, and \emph{Timeout} is when the verifier times out. We find that none of the Pensieve models satisfy the overall conjunctive specification, as some of the constituent specifications can be falsified. This indicates that the models, while optimizing for average reward, may not be trustworthy for practical usage. Moreover, we see that the SOTA verifier times out for some of the specifications for both Pensieve and Aurora. This suggests that the specifications are challenging for contemporary verifiers and can be used to guide the design of customized verifiers for NNs in computer systems. 

    {\begin{table}[tb]
        \centering
            
            \begin{tabular}{@{}lrrr@{}} 
                \toprule
                    Model & Verified & Falsified & Timeout \\
                  \midrule
                    Pensieve (small) & $5$ & $20$ & $5$\\
                    Pensieve (mid) & $2$ & $22$ & $6$\\
                    Pensieve (big) & $1$ & $28$ & $1$\\
                    \midrule 
                    Aurora (small) & $0$ & $0$ & $1$\\
                    Aurora (mid) & $0$ & $0$ & $1$\\
                    Original Aurora & $0$ & $0$ & $1$\\
                \bottomrule
            \end{tabular}
            \caption{Verifying \tool{}'s specifications sets consisting of $30$ specifications for ABR and $1$ specification for CC.}
            \label{tab:nnverification}
        \end{table}
    }
\subsubsection{Targeted attacks on NNs}
Next, we attack Pensieve to falsify \tool{}'s specification and study the attacks qualitatively. We generate Projected Gradient Descent attacks~\citep{pgd} on the models to identify inputs that cause the models to give extreme outputs (lowest/highest bit rates). As \tool{}'s overall specifications set is a conjunction of specifications, violating a single specification will falsify the specifications set. Hence, we attack the specifications from the generated specifications set that does not allow the extreme outputs in their postconditions. We show only the features of the attack input specified by the specifications. These features are sufficient to show the violation of intuitive behavior from the models. We show attacks on the Pensieve (big) model and note that similar attacks exist for other models too.

\begin{code}
\begin{center}
\begin{minipage}[t]{0.2\textwidth}
\centering
\small{\begin{align*}
    &\textbf{\text{Input}}:\\
    & BS = 11.2,\\
    & DT[-1] = 6.9,\\
    & DT[-2] = 2.9,\\
    & DT[-3] = 2.0\\
    &\textbf{\text{Output}}: BR = 300
\end{align*}}
\subcaption{}
\end{minipage}
\begin{minipage}[t]{0.2\textwidth}
\centering
\small{\begin{align*}
    &\textbf{\text{Input}}:\\
    & BS = 4.0,\\
    & DT[-1] = 11.8,\\
    & DT[-2] = 0.2,\\
    & DT[-3] = 0.2\\
    &\textbf{\text{Output}}: BR = 4300
\end{align*}}
\subcaption{}
\end{minipage}
\end{center}
\end{code}

Attack (a) consists of an input where the BS is high and only the last DT is high, with the other DTs low. The model still conservatively predicts the lowest bit rate, while a higher bit rate could be supported by the system. The Buffer-based (BB) algorithm, a simple ABR algorithm, can also predict a higher bitrate ($1850$ kbps) for this case. Attack (b), on the other hand, consists of an input where the buffer consists of only $1$ video chunk and the previous DT had been high. For this input, the model predicts the highest bit rate, which may result in rebuffering of the system and, therefore, affect the quality of experience for the users. The simple BB reference algorithm predicts $300$ kbps for this instance.

%% file: relatedwork.tex
\section{Related Work}


\textbf{Specifications for neural networks}. The current approach to generate specifications for neural networks (NNs) is largely dependent on human design.
Many existing works, such as \citep{verifying_learning_augmented_systems,wu2022scalable,wei2023building}, rely on experts to design their specifications.
Also, in the International Verification of Neural Networks Competition,
VNN-Comp~\citep{vnncomp23}, expert-designed specifications are used in benchmarks, including for Adaptive Bit Rate and Congestion Control.
However, the quality and relevance of these expert-designed
specifications remain unclear.
Recent work~\citep{geng2023towards} proposes to automatically mine neural activation patterns (NAP) as specifications.
NAP refers to the pattern of activation functions---whether they are activated
or deactivated---given a specific neural network and an input.
~\citet{geng2024learning} introduces multiple approaches to mine NAPs
for a given neural network. \tool{} differs from NAP mining as \tool{}'s specifications can generalize beyond the target neural networks and can be intuitively validated with domain knowledge. 

\textbf{Specification generation for programs}. There is a long history of work focused on synthesizing specifications for traditional programs, which has inspired \tool{}.
Unlike prior work~\citep{daikon,mining_specs,synthesizingspecs,precond_syn,synthesizing_contracts},
\tool{} targets neural networks instead of traditional general
programs. To the best of our knowledge, \tool{} is the first to mine
specifications for neural networks in computer systems using reference algorithms. \citet{perceptioncontract} also synthesize contracts for neural networks, but they operate in a setting with query-access to an oracle, which is not practically extensible to the applications we study. 

\textbf{NN verification}. NN verification formally verifies given neural networks for desirable properties such as robustness to input perturbations. It can be broadly classified as complete~\citep{bab1,bab2,bab3} and incomplete~\citep{abcrown,deeppoly} verification. NN verification is NP-complete~\citep{katz2017reluplexefficientsmtsolver}, which is hard to scale to larger NNs. However, the NNs in computer systems are generally small due to efficiency requirements and hence are conducive to verification. 


%% file: conclusion.tex
\section{Conclusion}
We present an automated approach for generating specifications for neural networks in applications where trustworthy reference algorithms exist. We formalize specification generation as an optimization problem and propose an effective algorithm \tool{}. We show specifications for two important applications --- adaptive bit rate setting and congestion control. We analyze the quality of \tool{}'s specifications and use them to verify and identify previously unknown vulnerabilities in SOTA neural networks.
%

%% file: appendix.tex
\newpage
\section{Pensieve's architectural details}\label{app:arch}
Pensieve's original architecture~\citep{pensieve}, that corresponds to our mid model has the following structure:

\textit{First Layer}: 3 parallel fully connected layer, each contains 128 neurons, and an 1D convolution layer with 128 filters and kernel size  is 4. These 4 layers take the input features in parallel.\\
\textit{Second Layer}: A fully connected (linear) layer with 128 neurons.\\
\textit{Output Layer}: A fully connected (linear) layer with 6 neurons.\\

Following the NN4Sys benchmarks in VNN Comp 2024 (\url{https://sites.google.com/view/vnn2024}), we also include the small and big models for Pensieve, having the following architectures. 

\textbf{Small}\\
\textit{First Layer}: 4 parallel fully connected layer, each contains 128 neurons.\\
\textit{Second Layer}: A fully connected (linear) layer with 128 neurons.\\
\textit{Output Layer}: fully connected (linear) layer with 6 neurons.\\

\textbf{Big}\\
\textit{First Layer}: 3 parallel fully connected layer, each contains 128 neurons, and an 1D convolution layer with 128 filters and kernel size  is 4. These 4 layers are parallel.\\
\textit{Second Layer}: A fully connected (linear) layer with 256 neurons.\\
\textit{Output Layer}: fully connected (linear) layer with 6 neurons.

\section{Training Aurora}\label{app:trainingaurora}


\textbf{Aurora model architectures.}
In this paper, we provide two different architectures for the Aurora model: the \textit{small model} and the \textit{mid model}. The \textit{mid model} retains the same architecture as the initial Aurora policy agent from the original paper~\citep{aurora}, which utilized a fully-connected neural network with two hidden layers of 32 $\rightarrow$ 16 neurons and employed a \textit{tanh} nonlinearity function. We have also developed a \textit{small model} with a similar architecture but scaled down to two hidden layers of 16 $\rightarrow$ 8 neurons, also using the \textit{tanh} nonlinearity.

\textbf{Training and testing setting.} The original Aurora model was trained in a gym simulation environment designed to replicate network links, with bandwidth and latency randomized to reflect real-world conditions.

To adapt Aurora for conditions similar to those experienced by BBR and Cubic, we made slight modifications to the simulation. This included using varied bandwidth from Pantheon traces (11 in total, available at \href{https://github.com/StanfordSNR/pantheon-traces}{Pantheon Traces}), as well as adjusting loss rate, packet queue size, and one-way delay. To ensure broad coverage of bandwidth scenarios, we increased the training steps.

In training, we simulate network conditions using Pantheon traces. Each condition includes:
\begin{itemize}
    \item \textbf{Bandwidth Trace}: Patterns of bandwidth over time.
    \item \textbf{Loss Rate}: Percentage of packets dropped.
    \item \textbf{Delay}: Time for a packet to travel one way.
    \item \textbf{Queue Size}: Maximum packets held in the network buffer before forwarding or dropping.
\end{itemize}

Using 11 traces, we had 18 distinct network conditions by varying loss, delay, and queue size. We split these conditions, with 75\% used for training and 25\% for testing, ensuring the RL-based Aurora model is not exposed to test patterns during training. During training and testing, we ensure that all testing network conditions are run at least once.

In our experiment, with a fixed random seed of 0, the split is as follows:
\begin{itemize}
    \item \textbf{Training Conditions}: 13, 14, 2, 5, 9, 8, 7, 15, 18, 6, 4, 11, 16
    \item \textbf{Testing Conditions}: 1, 3, 10, 12, 17
\end{itemize}

Network condition details can refer below:
\begin{table}[H]
    \centering
    \begin{tabular}{@{}c l c c c@{}}
        \toprule
        No. & Trace File & Delay & Loss & Queue Size \\
        \midrule
        1 & 0.57mbps-poisson & 28 & 0.0477 & 14 \\
        2 & 2.64mbps-poisson & 88 & 0 & 130 \\
        3 & 3.04mbps-poisson & 130 & 0 & 426 \\
        4 & 100.42mbps & 27 & 0 & 173 \\
        5 & 77.72mbps & 51 & 0 & 94 \\
        6 & 114.68mbps & 45 & 0 & 450 \\
        7 & 12mbps & 10 & 0 & 1 \\
        8 & 60mbps & 10 & 0 & 1 \\
        9 & 108mbps & 10 & 0 & 1 \\
        10 & 12mbps & 50 & 0 & 1 \\
        11 & 60mbps & 50 & 0 & 1 \\
        12 & 108mbps & 50 & 0 & 1 \\
        13 & 0.12mbps & 10 & 0 & 10000 \\
        14 & 10-every-200 & 10 & 0 & 10000 \\
        15 & 12mbps & 30 & 0 & 6 \\
        16 & 12mbps & 30 & 0 & 20 \\
        17 & 12mbps & 30 & 0 & 30 \\
        18 & 12mbps & 30 & 0 & 60 \\
        \bottomrule
    \end{tabular}
    \caption{Mapping of Network Parameters to Trace Files}
    \label{tab:network_params}
\end{table}

More details can be found in our network simulation implementation.

\textbf{Aurora Model Performance.} 
We evaluate the Aurora model on our testing network conditions. To benchmark its performance, we include a random model, which is randomly initialized and untrained. The \textit{Original} model refers to models trained in Aurora's original RL environment, while \textit{Current} refers to the model we trained under network conditions similar to BBR and Cubic, using Pantheon traces, which we introduced above.

{\begin{table}[H]
    \centering
        \begin{tabular}{@{}lrrr@{}} 
            \toprule
                Model & Random & Original & Current \\
              \midrule

                Aurora (small) & $631.38$ & $3195.15$ & $3279.03$\\
                Aurora (mid) & $631.38$ & $3273.63$ & $1380.22$\\
            \bottomrule
        \end{tabular}
        \caption{Aurora Model Rewards}
        \label{tab:verification}
    \end{table}
}


\section{DBSCAN clustering settings}\label{app:dbscan}
DBSCAN operates by identifying \emph{core points} in given data. Core points have a prespecified minimum number of points in their neighborhood, specified by a given radius $r$. We set the minimum number of samples $min_s$ to the number of points that make the cluster achieve the representation threshold $\threshrep$. For a low volume of specifications, we keep $r$ as the minimum radius that can ideally contain the minimum number of points, if densely packed. 

\section{Ablations}\label{sec:ablation}
\begin{figure*}[!htb]
    \centering
    \begin{subfigure}{\textwidth}
         \centering
         \includegraphics[width=\linewidth]{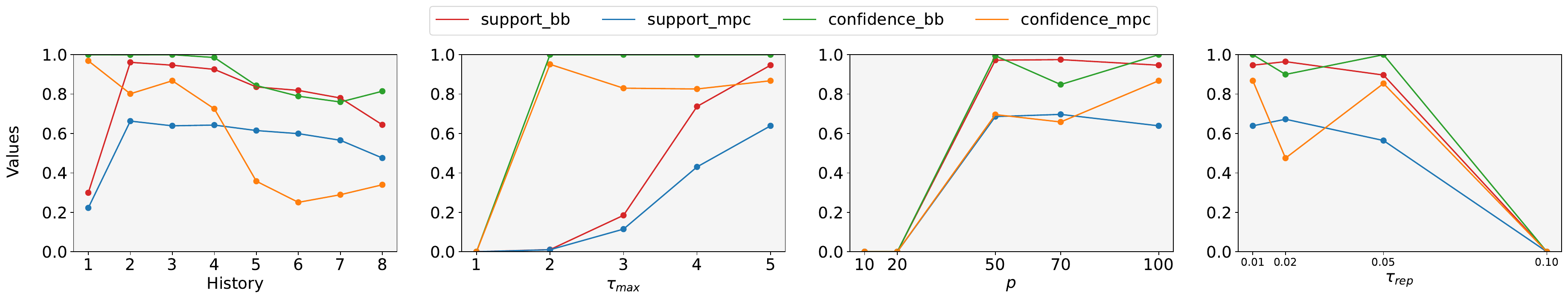}
         \caption{ABR}\label{Fig:ablationABR}
    \end{subfigure}
    \vspace{1em}
    \begin{subfigure}{\textwidth}
         \centering
         \includegraphics[width=\linewidth]{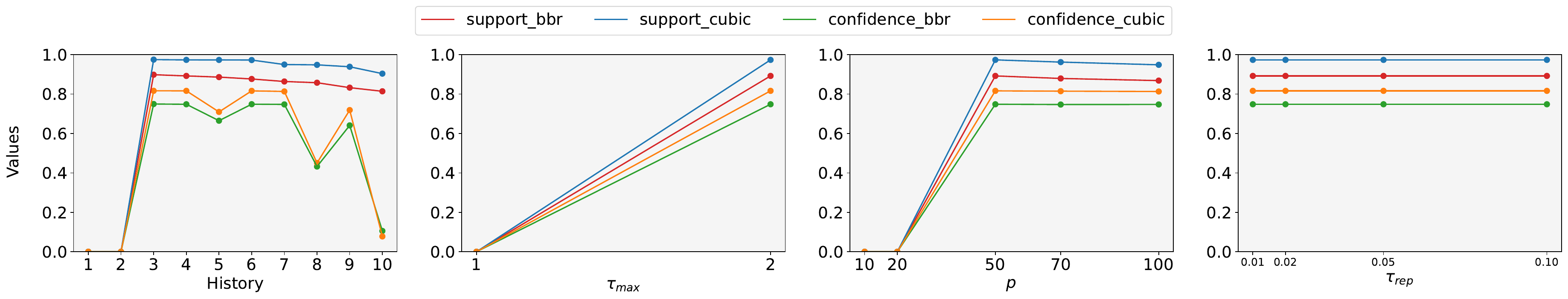}
         \caption{CC}\label{Fig:ablationCC}
    \end{subfigure}
\caption{Ablation study for hyperparameters affecting the specifications}
\end{figure*}

To select the best thresholds and \tool{}'s parameters, we study the variation in the support and confidence of \tool{}'s specifications over the training observations with the various settings. Specifically, we consider the history length of the features used in the specifications, representation threshold $\threshrep$, the number of partitions ($\parts$) of the input space $\inputs$ along each dimension, and the maximum permissible number of outputs in each specification $\threshtight$. Figures~\ref{Fig:ablationABR} and \ref{Fig:ablationCC} show the quality of the specifications for ABR and CC respectively. We select those parameters for our main experiments that yield specifications with high support and confidence over all the references, as observed in this ablation study. We select history = $3$ (previous $3$ download times will be used in specifications), $\threshrep=0.01, \parts=100, \threshtight=5$ for ABR and history = $4$ (previous $4$ observed features will be used in specifications), $\threshrep=0.01, \parts=50, \threshtight=2$ for CC. 

\section{\tool{}'s generated specification set for ABR}\label{app:specsets}
The following specifications~\ref{spec:abr_full} were generated by \tool{} for ABR, to be used in conjunction.

\addtocounter{listin}{-1}
\begin{code}
\captionof{listin}{Conjunctive specifications for ABR. (BS: Buffer Size, DT[$-i$]: $i^{th}$ last download time, BR: Bit Rate)}
\vspace{-1.5cm}
\begin{center}
\begin{minipage}{0.2\textwidth}
\centering
\small{\begin{align*}
&\textbf{\text{Precondition}}\\
& BS\in[0.4, 0.5],\\
& DT[-1]\in[0.15, 0.66],\\
& DT[-2]\in[0.15, 0.79],\\
& DT[-3]\in[0.54, 1.05]\\
&\textbf{\text{Postcondition}}\\
& BR\in\{300.0, 750.0, 1200.0, 2850.0\}
\end{align*}}
\subcaption*{1}
\end{minipage}
\begin{minipage}{0.2\textwidth}
\centering
\small{\begin{align*}
&\textbf{\text{Precondition}}\\
& BS\in[0.4, 0.5],\\
& DT[-1]\in[0.15, 1.05],\\
& DT[-2]\in[0.02, 0.79],\\
& DT[-3]\in[0.15, 1.05]\\
&\textbf{\text{Postcondition}}\\
& BR\in\{300.0, 750.0, 1200.0, 2850.0, 4300.0\}
\end{align*}}
\subcaption*{2}
\end{minipage}
\begin{minipage}{0.2\textwidth}
\centering
\small{\begin{align*}
&\textbf{\text{Precondition}}\\
& BS\in[0.4, 0.5],\\
& DT[-1]\in[0.28, 0.66],\\
& DT[-2]\in[0.15, 0.66],\\
& DT[-3]\in[0.54, 0.92]\\
&\textbf{\text{Postcondition}}\\
& BR\in\{300.0, 750.0, 2850.0\}
\end{align*}}
\subcaption*{3}
\end{minipage}
\begin{minipage}{0.2\textwidth}
\centering
\small{\begin{align*}
&\textbf{\text{Precondition}}\\
& BS\in[0.4, 0.5],\\
& DT[-1]\in[0.28, 0.66],\\
& DT[-2]\in[0.15, 0.79],\\
& DT[-3]\in[0.41, 1.05]\\
&\textbf{\text{Postcondition}}\\
& BR\in\{300.0, 750.0, 1200.0, 4300.0\}
\end{align*}}
\subcaption*{4}
\end{minipage}
\begin{minipage}{0.2\textwidth}
\centering
\small{\begin{align*}
&\textbf{\text{Precondition}}\\
& BS\in[0.4, 0.5],\\
& DT[-1]\in[0.28, 0.66],\\
& DT[-2]\in[0.28, 0.66],\\
& DT[-3]\in[0.54, 0.92]\\
&\textbf{\text{Postcondition}}\\
& BR\in\{300.0, 750.0\}
\end{align*}}
\subcaption*{5}
\end{minipage}
\begin{minipage}{0.2\textwidth}
\centering
\small{\begin{align*}
&\textbf{\text{Precondition}}\\
& BS\in[0.4, 0.5],\\
& DT[-1]\in[0.28, 0.66],\\
& DT[-2]\in[0.54, 1.05],\\
& DT[-3]\in[0.28, 0.66]\\
&\textbf{\text{Postcondition}}\\
& BR\in\{300.0, 750.0, 1200.0, 2850.0, 4300.0\}
\end{align*}}
\subcaption*{6}
\end{minipage}
\begin{minipage}{0.2\textwidth}
\centering
\small{\begin{align*}
&\textbf{\text{Precondition}}\\
& BS\in[0.4, 0.5],\\
& DT[-1]\in[0.28, 0.66],\\
& DT[-2]\in[0.66, 1.05],\\
& DT[-3]\in[0.28, 0.66]\\
&\textbf{\text{Postcondition}}\\
& BR\in\{300.0, 750.0, 1200.0\}
\end{align*}}
\subcaption*{7}
\end{minipage}
\begin{minipage}{0.2\textwidth}
\centering
\small{\begin{align*}
&\textbf{\text{Precondition}}\\
& BS\in[0.4, 0.5],\\
& DT[-1]\in[0.28, 1.05],\\
& DT[-2]\in[0.15, 0.79],\\
& DT[-3]\in[0.15, 0.66]\\
&\textbf{\text{Postcondition}}\\
& BR\in\{300.0, 750.0, 1200.0, 2850.0\}
\end{align*}}
\subcaption*{8}
\end{minipage}
\begin{minipage}{0.2\textwidth}
\centering
\small{\begin{align*}
&\textbf{\text{Precondition}}\\
& BS\in[0.4, 0.5],\\
& DT[-1]\in[0.54, 1.05],\\
& DT[-2]\in[0.15, 0.66],\\
& DT[-3]\in[0.15, 0.54]\\
&\textbf{\text{Postcondition}}\\
& BR\in\{300.0, 750.0, 1850.0\}
\end{align*}}
\subcaption*{9}
\end{minipage}
\begin{minipage}{0.2\textwidth}
\centering
\small{\begin{align*}
&\textbf{\text{Precondition}}\\
& BS\in[0.4, 0.5],\\
& DT[-1]\in[0.54, 1.05],\\
& DT[-2]\in[0.15, 0.66],\\
& DT[-3]\in[0.15, 0.66]\\
&\textbf{\text{Postcondition}}\\
& BR\in\{300.0, 750.0, 1200.0\}
\end{align*}}
\subcaption*{10}
\end{minipage}
\begin{minipage}{0.2\textwidth}
\centering
\small{\begin{align*}
&\textbf{\text{Precondition}}\\
& BS\in[0.4, 0.54],\\
& DT[-1]\in[0.15, 0.66],\\
& DT[-2]\in[0.15, 0.79],\\
& DT[-3]\in[0.41, 1.05]\\
&\textbf{\text{Postcondition}}\\
& BR\in\{300.0, 750.0, 1850.0, 2850.0\}
\end{align*}}
\subcaption*{11}
\end{minipage}
\begin{minipage}{0.2\textwidth}
\centering
\small{\begin{align*}
&\textbf{\text{Precondition}}\\
& BS\in[0.4, 0.54],\\
& DT[-1]\in[0.54, 1.05],\\
& DT[-2]\in[0.15, 0.66],\\
& DT[-3]\in[0.15, 0.66]\\
&\textbf{\text{Postcondition}}\\
& BR\in\{300.0, 750.0, 1850.0, 2850.0, 4300.0\}
\end{align*}}
\subcaption*{12}
\end{minipage}
\begin{minipage}{0.2\textwidth}
\centering
\small{\begin{align*}
&\textbf{\text{Precondition}}\\
& BS\in[0.4, 1.09],\\
& DT[-1]\in[0.02, 1.18],\\
& DT[-2]\in[0.02, 1.18],\\
& DT[-3]\in[0.02, 1.18]\\
&\textbf{\text{Postcondition}}\\
& BR\in\{300.0, 750.0, 1200.0, 1850.0, 4300.0\}
\end{align*}}
\subcaption*{13}
\end{minipage}
\begin{minipage}{0.2\textwidth}
\centering
\small{\begin{align*}
&\textbf{\text{Precondition}}\\
& BS\in[0.4, 1.48],\\
& DT[-1]\in[0.02, 1.18],\\
& DT[-2]\in[0.02, 1.18],\\
& DT[-3]\in[0.02, 1.18]\\
&\textbf{\text{Postcondition}}\\
& BR\in\{300.0, 750.0, 1200.0, 1850.0, 2850.0\}
\end{align*}}
\subcaption*{14}
\end{minipage}
\begin{minipage}{0.2\textwidth}
\centering
\small{\begin{align*}
&\textbf{\text{Precondition}}\\
& BS\in[0.5, 0.61],\\
& DT[-1]\in[0.54, 0.92],\\
& DT[-2]\in[0.15, 0.54],\\
& DT[-3]\in[0.15, 0.54]\\
&\textbf{\text{Postcondition}}\\
& BR\in\{300.0, 750.0, 1200.0, 2850.0, 4300.0\}
\end{align*}}
\subcaption*{15}
\end{minipage}
\begin{minipage}{0.2\textwidth}
\centering
\small{\begin{align*}
&\textbf{\text{Precondition}}\\
& BS\in[0.61, 0.75],\\
& DT[-1]\in[0.02, 0.41],\\
& DT[-2]\in[0.41, 1.05],\\
& DT[-3]\in[0.15, 0.66]\\
&\textbf{\text{Postcondition}}\\
& BR\in\{300.0, 750.0, 1850.0\}
\end{align*}}
\subcaption*{16}
\end{minipage}
\begin{minipage}{0.2\textwidth}
\centering
\small{\begin{align*}
&\textbf{\text{Precondition}}\\
& BS\in[0.61, 0.78],\\
& DT[-1]\in[0.02, 0.41],\\
& DT[-2]\in[0.41, 1.05],\\
& DT[-3]\in[0.15, 0.66]\\
&\textbf{\text{Postcondition}}\\
& BR\in\{300.0, 750.0, 1850.0, 4300.0\}
\end{align*}}
\subcaption*{17}
\end{minipage}
\begin{minipage}{0.2\textwidth}
\centering
\small{\begin{align*}
&\textbf{\text{Precondition}}\\
& BS\in[0.61, 0.78],\\
& DT[-1]\in[0.02, 0.41],\\
& DT[-2]\in[0.41, 1.18],\\
& DT[-3]\in[0.15, 0.66]\\
&\textbf{\text{Postcondition}}\\
& BR\in\{300.0, 750.0, 1200.0, 4300.0\}
\end{align*}}
\subcaption*{18}
\end{minipage}
\begin{minipage}{0.2\textwidth}
\centering
\small{\begin{align*}
&\textbf{\text{Precondition}}\\
& BS\in[0.61, 0.82],\\
& DT[-1]\in[0.02, 0.41],\\
& DT[-2]\in[0.41, 1.05],\\
& DT[-3]\in[0.15, 0.66]\\
&\textbf{\text{Postcondition}}\\
& BR\in\{300.0, 750.0, 1850.0, 2850.0, 4300.0\}
\end{align*}}
\subcaption*{19}
\end{minipage}
\begin{minipage}{0.2\textwidth}
\centering
\small{\begin{align*}
&\textbf{\text{Precondition}}\\
& BS\in[0.61, 1.06],\\
& DT[-1]\in[0.02, 0.41],\\
& DT[-2]\in[0.02, 1.18],\\
& DT[-3]\in[0.15, 1.18]\\
&\textbf{\text{Postcondition}}\\
& BR\in\{300.0, 750.0, 1200.0, 2850.0, 4300.0\}
\end{align*}}
\subcaption*{20}
\end{minipage}
\begin{minipage}{0.2\textwidth}
\centering
\small{\begin{align*}
&\textbf{\text{Precondition}}\\
& BS\in[0.64, 0.75],\\
& DT[-1]\in[0.02, 0.41],\\
& DT[-2]\in[0.15, 0.54],\\
& DT[-3]\in[0.54, 1.18]\\
&\textbf{\text{Postcondition}}\\
& BR\in\{300.0, 750.0, 1200.0\}
\end{align*}}
\subcaption*{21}
\end{minipage}
\begin{minipage}{0.2\textwidth}
\centering
\small{\begin{align*}
&\textbf{\text{Precondition}}\\
& BS\in[0.64, 0.78],\\
& DT[-1]\in[0.02, 0.41],\\
& DT[-2]\in[0.15, 0.54],\\
& DT[-3]\in[0.54, 0.92]\\
&\textbf{\text{Postcondition}}\\
& BR\in\{300.0, 750.0, 1850.0, 2850.0\}
\end{align*}}
\subcaption*{22}
\end{minipage}
\begin{minipage}{0.2\textwidth}
\centering
\small{\begin{align*}
&\textbf{\text{Precondition}}\\
& BS\in[0.75, 0.82],\\
& DT[-1]\in[0.02, 0.41],\\
& DT[-2]\in[0.15, 0.54],\\
& DT[-3]\in[0.54, 0.92]\\
&\textbf{\text{Postcondition}}\\
& BR\in\{300.0, 750.0, 1200.0\}
\end{align*}}
\subcaption*{23}
\end{minipage}
\begin{minipage}{0.2\textwidth}
\centering
\small{\begin{align*}
&\textbf{\text{Precondition}}\\
& BS\in[0.78, 0.92],\\
& DT[-1]\in[0.02, 0.41],\\
& DT[-2]\in[0.54, 0.92],\\
& DT[-3]\in[0.28, 0.66]\\
&\textbf{\text{Postcondition}}\\
& BR\in\{300.0, 750.0, 1200.0\}
\end{align*}}
\subcaption*{24}
\end{minipage}
\begin{minipage}{0.2\textwidth}
\centering
\small{\begin{align*}
&\textbf{\text{Precondition}}\\
& BS\in[0.82, 0.89],\\
& DT[-1]\in[0.02, 0.41],\\
& DT[-2]\in[0.54, 0.92],\\
& DT[-3]\in[0.28, 0.66]\\
&\textbf{\text{Postcondition}}\\
& BR\in\{300.0, 750.0, 1850.0, 2850.0\}
\end{align*}}
\subcaption*{25}
\end{minipage}
\begin{minipage}{0.2\textwidth}
\centering
\small{\begin{align*}
&\textbf{\text{Precondition}}\\
& BS\in[0.82, 0.96],\\
& DT[-1]\in[0.02, 0.41],\\
& DT[-2]\in[0.41, 0.92],\\
& DT[-3]\in[0.15, 0.66]\\
&\textbf{\text{Postcondition}}\\
& BR\in\{300.0, 750.0, 1200.0, 2850.0\}
\end{align*}}
\subcaption*{26}
\end{minipage}
\begin{minipage}{0.2\textwidth}
\centering
\small{\begin{align*}
&\textbf{\text{Precondition}}\\
& BS\in[0.85, 0.96],\\
& DT[-1]\in[0.02, 0.41],\\
& DT[-2]\in[0.02, 0.41],\\
& DT[-3]\in[0.28, 0.66]\\
&\textbf{\text{Postcondition}}\\
& BR\in\{300.0, 750.0, 1200.0\}
\end{align*}}
\subcaption*{27}
\end{minipage}
\begin{minipage}{0.2\textwidth}
\centering
\small{\begin{align*}
&\textbf{\text{Precondition}}\\
& BS\in[0.85, 1.06],\\
& DT[-1]\in[0.02, 0.41],\\
& DT[-2]\in[0.02, 0.41],\\
& DT[-3]\in[0.54, 1.05]\\
&\textbf{\text{Postcondition}}\\
& BR\in\{300.0, 750.0, 1200.0\}
\end{align*}}
\subcaption*{28}
\end{minipage}
\begin{minipage}{0.2\textwidth}
\centering
\small{\begin{align*}
&\textbf{\text{Precondition}}\\
& BS\in[1.09, 1.2],\\
& DT[-1]\in[0.02, 0.54],\\
& DT[-2]\in[0.02, 0.54],\\
& DT[-3]\in[0.41, 0.79]\\
&\textbf{\text{Postcondition}}\\
& BR\in\{300.0, 750.0, 1200.0, 1850.0\}
\end{align*}}
\subcaption*{29}
\end{minipage}
\begin{minipage}{0.2\textwidth}
\centering
\small{\begin{align*}
&\textbf{\text{Precondition}}\\
& BS\in[1.09, 1.23],\\
& DT[-1]\in[0.41, 0.79],\\
& DT[-2]\in[0.15, 0.66],\\
& DT[-3]\in[0.15, 0.54]\\
&\textbf{\text{Postcondition}}\\
& BR\in\{750.0, 1200.0, 1850.0, 2850.0, 4300.0\}
\end{align*}}
\subcaption*{30}
\end{minipage}
\label{spec:abr_full}
\end{center}
\end{code}